\documentclass{ecai}
\usepackage{graphicx}
\usepackage{latexsym}
\usepackage{color}
\usepackage{makecell}
\usepackage{multirow}
\usepackage{amsmath}
\usepackage{amsfonts,amssymb}
\usepackage{booktabs}

\begin{document}

\begin{frontmatter}

\title{DPBERT: Efficient Inference for BERT based on Dynamic Planning}

\author[A]{\fnms{Weixin}~\snm{Wu}}
\author[A]{\fnms{Hankz Hankui}~\snm{Zhuo}\thanks{Corresponding Author. Email: zhuohank@mail.sysu.edu.cn. To appear in ECAI 2023.}}

\address[A]{Sun Yat-sen University, Guangzhou, China}

\begin{abstract}
Large-scale pre-trained language models such as BERT have contributed significantly to the development of NLP. However, those models require large computational resources, making it difficult to be applied to mobile devices where computing power is limited. In this paper we aim to address the weakness of existing input-adaptive inference methods which fail to take full advantage of the structure of BERT. We propose Dynamic Planning in BERT, a novel fine-tuning strategy that can accelerate the inference process of BERT through selecting a subsequence of transformer layers list of backbone as a computational path for an input sample. To do this, our approach adds a planning module to the original BERT model to determine whether a layer is included or bypassed during inference. Experimental results on the GLUE benchmark exhibit that our method reduces latency to 75\% while maintaining 98\% accuracy, yielding a better accuracy-speed trade-off compared to state-of-the-art input-adaptive methods.
\end{abstract}

\end{frontmatter}

\section{Introduction}
Recently, pre-training language models (PLMs) have shown powerful capabilities in Natural Language Processing (NLP), achieving sota-level results in a tremendous amount of tasks. Despite the promotion in accuracy, the number of parameters of those PLMs, such as BERT \cite{DBLP:conf/naacl/DevlinCLT19}, RoBERTa \cite{DBLP:journals/corr/abs-1907-11692} and XLNet \cite{DBLP:conf/nips/YangDYCSL19}, reaches millions or even billions, which renders them costly to do inference. This drawbacks make it even more challenging when we perform the inference on mobile devices due to sluggish computation speed. Therefore, it is desirable to minimize the inference time of the PLMs while maintaining an acceptable accuracy.

In order to deal with the above-mentioned issues, approaches have been proposed to accelerate the inference, such as early exiting \cite{DBLP:conf/acl/XinTLYL20}, knowledge distillation \cite{DBLP:journals/corr/RomeroBKCGB14}, and quantization \cite{shen2020q} pruning \cite{DBLP:journals/corr/HanPTD15}, among which early exiting is a reference acceleration method designed for models with repetitive architecture. Recently, early exiting was applied to the variants of BERT, which consist of a sequence of transformer layers \cite{DBLP:conf/nips/VaswaniSPUJGKP17} and a task-specific classifier. According to the degree of difficulty of tasks, early exiting could be performed on one of the intermediate classifiers through a specific exiting decision mechanism during inference. In this work we focus on designing a novel mechanism of early exiting.

From the observation that lower layers of PLMs may capture more syntactic information while higher layers may capture more high-level semantic information \cite{DBLP:conf/acl/TenneyDP19}, we conjecture that each transformer layer has different functionalities. Previous early exiting approaches reduce the computation of several of the top layers to achieve inference acceleration \cite{DBLP:conf/acl/SchwartzSSDS20,DBLP:conf/acl/XinTLYL20}. Due to the constraint of the mechanism, however, the calculation of the bottom and middle layers could not be bypassed during inference, which means the inference path (composed of different layers of the PLMs) of a task cannot be selected ``freely''. 

\begin{figure}[t]
    \centering
    \includegraphics[scale=0.5]{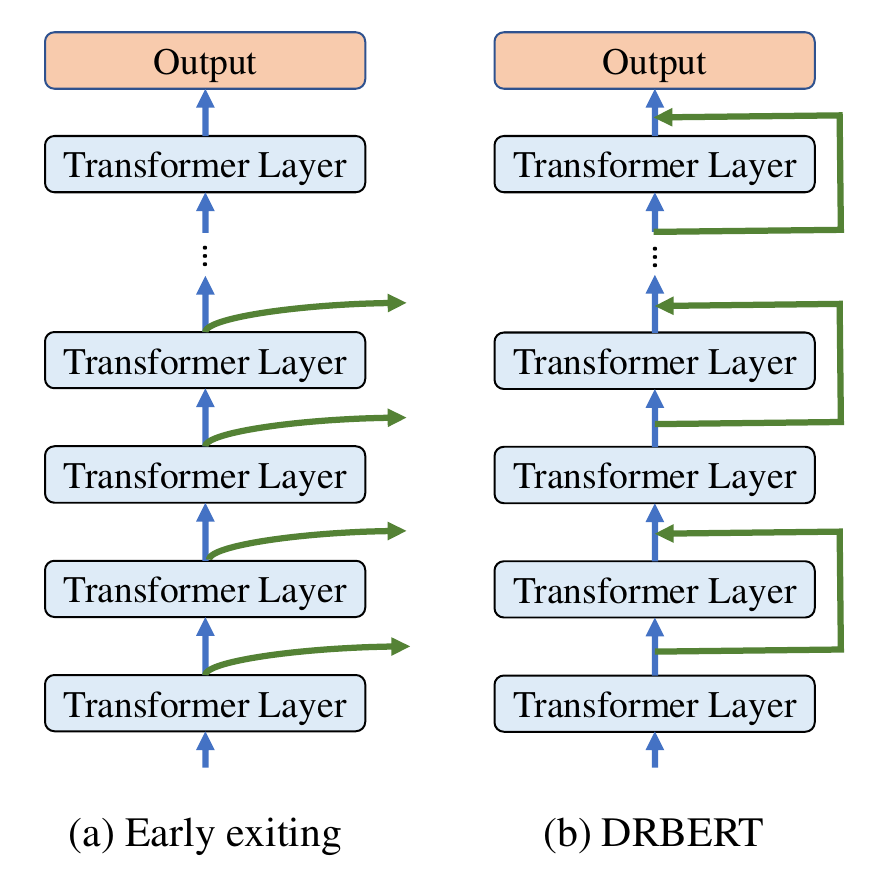}
    \caption{Comparison of inference between early exiting and dynamic planning.}\label{intro_compare}
\end{figure}

To address those issues, we propose {\tt DPBERT}, which stands for \textbf{D}ynamic \textbf{P}lanning for \textbf{BERT}, a novel framework that constructs an inference path with arbitrary transformer layers skipped. As described in \cite{DBLP:conf/eccv/WangYDDG18}, in the field of computer vision, deep representations are only necessary for a small percentage of images, and thus it can achieve model acceleration through adding a dynamic planning mechanism to ResNet  \cite{DBLP:conf/cvpr/HeZRS16}, i.e., it determines whether a layer of a convolutional neural network should be included when processing a given image. Inspired by the above work, we assume that it is also effective to apply dynamic planning mechanisms in BERT, since BERT consists of multiple repetitive networks similar to ResNet. Given the input text, we construct hidden states corresponding to the transformer layers, which determine the planning decision of the layers. After that, we build a planning policy network corresponding to each transformer layer to map hidden states to decision actions, i.e., skipping or executing the subsequent layer.

Figure \ref{intro_compare} compares the inference procedure between previous early exiting approaches and our {\tt DPBERT}. In previous early exiting, the inference process skips all of the layers that are after a layer being determined to be skipped (and directly calculates the classification or regression result based on the layers that are before the first layer being skipped). In comparison, our {\tt DPBERT} determines whether a transformer layer is included in the inference procedure, allowing various combination of transformer layers for inference. 
We conjecture that our {\tt DPBERT} possesses better trade-offs between model performance and inference time based on the flexible planning mechanism.

Learning an efficient planning strategy is critically and challenging. To compress the computation while maintaining accuracy, it is important to appropriately bypass the unnecessary transformer layers in BERT for each input sample. During inference, a sequence of transformer layers is decided whether to be included into the computational path or not. To solve this sequential decision problem, we add reinforcement learning to our framework, using token representation as state and planning networks as policy network. Through rewarding the actions of skipping layers but maintaining accuracy, our model learns to accomplish a better accuracy-speed trade-off. We experimentally demonstrate that the addition of reinforcement learning is more effective compared to soft approximations (Section \ref{Ablation Study}).

In summary, our contributions are as follows:

\begin{itemize}
    \item We present {\tt DPBERT}, a novel and universal input-adative inference mechanism. Compared with early exiting, it allows unrestricted choice of computational path for input samples.
    \item We propose a multi-stage training approach incorporating reinforcement learning that allows for a shorter computational path while preserving as much of the model's effectiveness as possible.
    \item Extensive and massive experiments are conducted on GLUE benchmark. Compared to BERT, {\tt DPBERT} can accelerate inference by up to 1.34$\times$ while maintaining 98\% accuracy.
\end{itemize}

\section{Related Work}
The existing mainstream inference acceleration methods for pre-trained models fall into two main types: (1) Static approaches reduce the computational effort of each sample inference by reducing the model parameters, i.e. there is no difference in the layers that each sample passes through. (2) Input-Adaptive approaches allow the model to select different computational paths based on instances during inference. Therefore, simpler input instances require shorter model paths to complete the inference. In this way, the computation time of model inference can be effectively reduced. The method we propose belongs to the second category mentioned above.

\subsection{Static Approaches}
There are various established techniques to speed-up model inference in the context of deep learning. Some of these methods have been shown to be effective in PLMs acceleration.

Knowledge Distillation \cite{hinton2015distilling} offers a practical way to transfer the knowledge stored in a teacher model to a lightweight student model, which is more computationally effective.  Pruning \cite{gordon2020compressing,michel2019sixteen} removes redundant parameters or unimportant modules such as attention heads and feed forward layers of the model. Quantization \cite{shen2020q,wu2016quantized} is a method to lowers the demand for the numbers of bits while running and storing a model. Matrix decomposition \cite{DBLP:conf/iclr/LanCGGSS20} decomposes the large vocabulary embedding matrix into two smaller matrices, making it easier to grow the hidden size without significantly increasing the size of the parameters size of the vocabulary embedding. These static approaches usually require pre-training the model from scratch. In these ways, all the input samples have to go through a fixed computational graph. In comparison, input-adaptive approaches can assign different computation path to samples.

\subsection{Input-Adaptive Approaches}
An alternative strategy for enhancing the efficiency of the model is to perform adaptive inference for various input samples. This method allocated fewer resources to certain parts of the input, thereby potentially reducing inference time. Adaptive Computation Time \cite{graves2016adaptive} presents a trainable halting mechanism to construct computational paths adaptively when inferring. Extended from it, early exiting is explored to apply in pre-training language models. Depending on different metrics or strategies, early exit allowed the model to stop the computation at any layer of the network during the inference phase and obtain the final result immediately through the corresponding additional classifier. RightTool \cite{schwartz2020right} uses the calibrated confidence scores of classifiers to make exit decisions. FastBERT \cite{liu2020fastbert} and DeeBERT \cite{DBLP:conf/acl/XinTLYL20} evaluated a classifier's confidence based on the entropy of the output probability distribution and decided to terminate when it exceeds a set threshold. BERxiT \cite{xin2021berxit} proposed to apply a learning-to-exit network module to make the decision and extended early exit to regression tasks. CascadeBERT \cite{DBLP:conf/emnlp/LiLCRLZS21} found that early exiting faces a performance bottleneck under high speed-up ratios and thus they proposed a new framework based on cascading mechanism.

However, early exiting does not allow arbitrary skipping of layers in the model due to its mechanism, i.e., the model can only save the computation of the top layers. In the field of computer vision, BlockDrop \cite{wu2018blockdrop} and SkipNet \cite{DBLP:conf/eccv/WangYDDG18} used reinforcement learning to determine the dynamic planning computation graph for the inference phase. In this way, the model was able to construct a specific computational graph for each sample without restrictions. Inspired by the recent work, we propose {\tt DPBERT}, a novel pre-training model acceleration framework that learning to construct a sample-specific computational path by discarding redundant transformer layers.

\section{Methodology}
\begin{figure*}[!ht]
    \centering
    \includegraphics[scale=0.45]{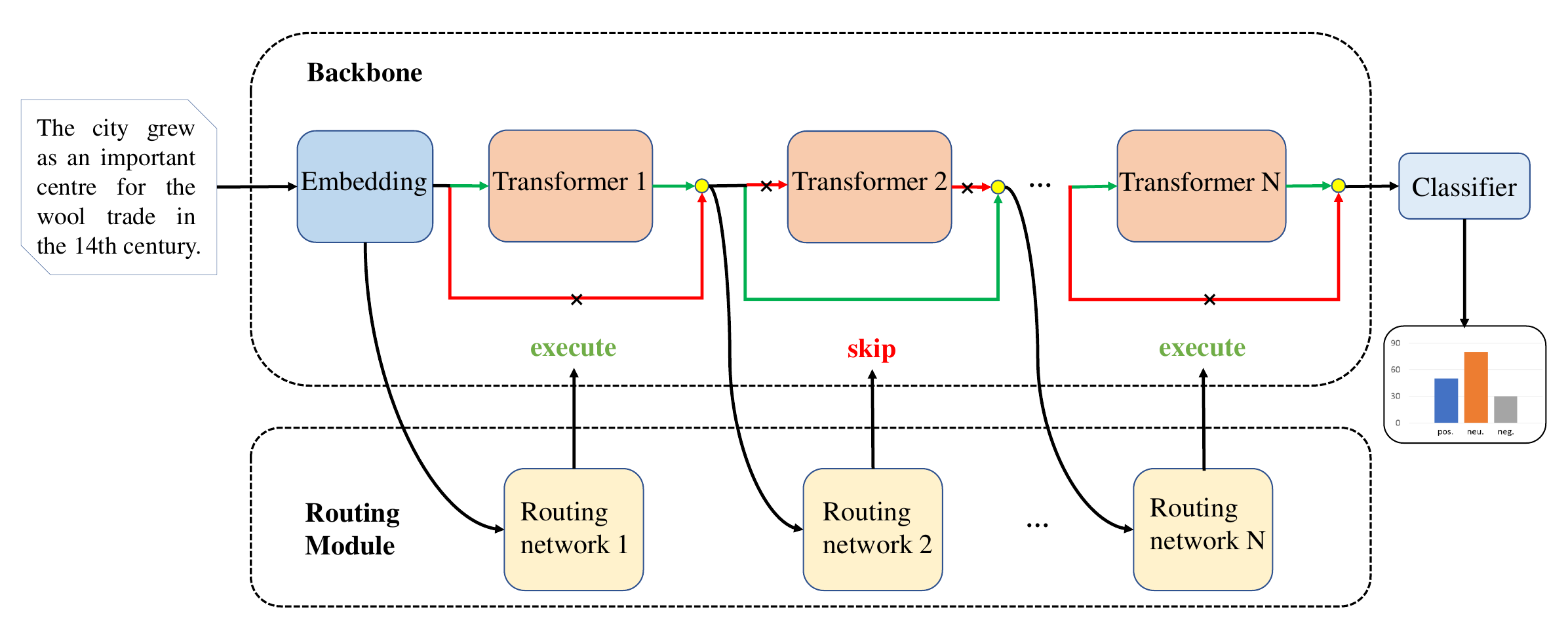}
    \caption{The inference process of {\tt DPBERT}, where the subsequence of transformer layers varies based on the complexity of instance. It shows a sequence of dynamic planning strategy. During inference, whether each transformer layer is executed or not is determined by the action output from the corresponding planning network. When the action is executing, the current layer is included in the computational path (marked in green in the figure), and when the action is skipping, the current layer is bypassed (marked in red in the figure).}\label{overview}
\end{figure*}

To address that early exiting can only skip the computation of the top transformer layers, we develop {\tt DPBERT}, which modifies fine-tuning and inference of BERT model with no change in pre-training. It adds one planning network to each transformer layer, which determines whether to execute the next layer. If the decision is to execute it, it will be computed normally as the original BERT. Otherwise, the output of the current transformer layer will skip the computation of the next transformer layer and be directly input to the subsequent network layer. An overview of {\tt DPBERT} is shown in Figure \ref{overview}.

\subsection{Model Architecture}
The model architecture of {\tt DPBERT} is shown in Figure \ref{overview}. We can see that {\tt DPBERT} consists of backbone and planning module. A 12-layers transformer encoder and an additional classifier compose the backbone model, while planning module is a sequence of planning networks correspond to transformer layers.

\subsubsection{Backbone Model}
The backbone consists of two parts: an embedding layer and a chain of transformer layers \cite{vaswani2017attention}. These structures are in line with those of BERT. A common inference process of BERT can be described as the input instance passes through transformer layers and classifiers to predict the result. 

In detail, given an input sentence $x=[w_0,w_1,...,w_n]$ of length $n$, the model will first embed it as vector representations $\boldsymbol{e}$ as \eqref{embed}:
\begin{equation}
    \boldsymbol{e} = \text{Embedding}(x) \label{embed}
\end{equation}
where $\boldsymbol{e}$ is the summation of word, position, and segment embeddings. Then the word representations pass will through a series of transformer layers to extract feature like \eqref{transformer_layer}:
\begin{equation}
    \boldsymbol{h^i} = \text{Trans}^i(\boldsymbol{h^{i-1}}) \label{transformer_layer}
\end{equation}
where $\text{Trans}^i$ is the $i^{th}$ transformer layer, $\boldsymbol{h^i} (i=0,1,...,L)$ represents the hidden state produced by $i^{th}$ layer, and $\boldsymbol{h^0}=\boldsymbol{e}$. The number of transformer layers is $L$.

Finally, the hidden vectors of the last layer goes through the output layer to predict the class label distribution or a value (for regression, assuming an output dimension of 1 for the output).

\subsubsection{Planning Module}
Based on the framework of the backbone model, we make changes to the part of transformer layers so that the model can freely construct the computational paths based on the input instances during inference.
We propose to add a corresponding lightweight planning network $\text{Plan}^i$ to decide whether to enter the layer $\text{Trans}^{i}$. In this case, the number of calculation paths that can be selected by the model increases to $2^L$, indicating that the samples of different difficulty could pass through paths of various lengths. We input the hidden state of the $i^{th}$ layer to the corresponding planning network and makes a decision whether the next layer to be executed or not.
\begin{equation}
    a^{i} = \text{Plan}^i(\boldsymbol{h^{i-1}}) \label{routeNet}
\end{equation}
Where $a^i$ is a binary action. When $a^i=1$ means that the $i^{th}$ transformer layer will be executed and vice versa. To avoid yielding too complex calculations, we apply a lightweight fully connected network as then planning network and  expect it to make the decisions for the subsequent transformer layer based on the previous hidden state. However, if the hidden state of all tokens is applied as input, the planning network will be still so computationally intensive that inference acceleration cannot be achieved. Therefore, we only select the hidden vector corresponding to some of these tokens as the state for decision-making. 

In the original BERT model, a single-layer fully-connected network is attached to the final transformer layer as the classifier, and then the classifier is updated with backbone jointly. Finally, The classifier outputs the prediction which is derived from the vector corresponding to the CLS token and the other word-vectors are ignored. This process gradually converges the information to the hidden state corresponding to the token CLS, which means that it contains a wealth of classification knowledge. Therefore, in this paper, we only regard the hidden state corresponding to CLS as the priori knowledge for planning module. Concretely, the planning network is a lightweight one-layer fully-connected network which takes the hidden state $\boldsymbol{h^i_{\text{CLS}}}$ as input  and provides the decision action $a^i$ at the $i^{th}$ layer:
\begin{equation}
    s^i = \text{sigmoid}(\boldsymbol{h^{i-1}_{\text{CLS}}}\boldsymbol{A^T} + b)  \label{fc} 
\end{equation}
where $\boldsymbol{A}$ and $b$ are both learnable parameter, which are not shared between different planning networks. We map the output of the planning network as a binary action like \eqref{action}:
\begin{equation}
    a^i = \mathbb{I}(s^i>0.5) \label{action}
\end{equation}
Depending on the decision action, the model will choose to execute or bypass the next layer. The process is as follows:
\begin{equation}
    \boldsymbol{h^{i}} = a^i\text{Trans}^i(\boldsymbol{h^{i-1}}) + (1-a^i)\boldsymbol{h^{i-1}}    
\end{equation}
In contrast to the original BERT model where the hidden state has to go through all layers, the $\boldsymbol{h^i}$ in {\tt DPBERT}  is only passed a subsequence of the backbone model guided by the planning module. In order to balance model performance and inference latency, it is critical for planning networks to identify layers that redundant to a sample. Consequently, we elaborate how the planning network learn to choose the inference computation path through reinforcement learning in Section \ref{Model Training}.

\subsection{Model Training} \label{Model Training}
For downstream inference, {\tt DPBERT} requires three training steps: backbone fine-tuning, the planning networks initialization and reinforcement learning for the entire model. In the first stage, we use the original method to fine-tune the backbone model. Then, we add the planning module and train it with the parameters in the backbone frozen. In the last stage, the backbone and the planning module need to be updated jointly in the second stage. 

\subsubsection{Fine-tuning for Backbone}
For the downstream tasks, we take as input the task-specific data to fine-tune the backbone model. At this stage, the planning module is not yet operational and the inference procedure of the transformer layer is the same with the original BERT model \cite{DBLP:conf/naacl/DevlinCLT19}.

\subsubsection{Initialization for Planning Module} \label{intialization}
In the second stage, we need to tune planning module because the parameters of the planning networks are initialized at random.  To compute the gradient, we relax the output of the planning network to a continuous range of [0,1]. In other words, the sigmoid results are directly utilized as the output of the planning networks, i.e. $a^i=s^i$, thus allowing for gradient back propagation. 
With the parameter of the backbone frozen, We perform supervised training with task-specific labelled data. This approach can effectively initialize the parameters of the planning network. Subsequently, we jointly train the backbone and planning module to enable the model to discard redundant layers while maintaining excellent ability.

\begin{table*}[t]
	\centering

		\resizebox{0.95\textwidth}{!}{
	\begin{tabular}{l|c|cccccccc|c} 
		\toprule
		Model &speed-up &\makecell[c]{CoLA\\(8.5K)} &\makecell[c]{MNLI-(m/mm)\\(393K)} &\makecell[c]{SST-2\\(67K)} &\makecell[c]{QNLI\\(105K)} &\makecell[c]{MRPC\\(3.7K)} &\makecell[c]{QQP\\(364K)} &\makecell[c]{RTE\\(2.5K)} &\makecell[c]{STS-B\\(5.7K)} &Avg\\ 
		\hline
		\multicolumn{10}{c}{\textit{Dev.Set}} \\
            \hline
		$\text{BERT-base}$ &1.00$\times$ &57.2 &84.7 &92.1 &91.3 &88.1 &89.8 &67.9 &88.3 &82.4 \\
            \hline
            Layer Drop &1.94$\times$ &45.4 &80.7 &90.7 &88.4 &85.9 &88.3 &65.2 &85.7 &78.8 \\
            DistilBERT &1.63$\times$ &51.3 &82.2 &
            \underline{\pmb{91.3}} &\underline{\pmb{89.2}} &87.5 &88.5 &59.9 &86.9 &79.6 \\
            \hline
            DEEBERT &1.33$\times$ &41.2 &79.2  &90.5 &87.0 &82.9 &86.4 &64.6 &- &76.0 \\
            BERxiT &1.40$\times$ &42.0 &81.9 &90.7 &88.0 &81.8 &88.9 &63.2 &87.7 &78.0 \\
            PaBEE &1.33$\times$ &51.1 &82.1 &90.6 &87.9 &82.6 &\pmb{89.0} &65.7 &88.7 &79.7 \\
            DPBERT(ours) &1.34$\times$ &\pmb{56.8} &\pmb{82.5} &\pmb{90.8} &\pmb{88.7} &\pmb{87.8} &88.3 &\pmb{66.8} &\pmb{88.7} &\pmb{81.3} \\
            \hline
            \hline
		\multicolumn{10}{c}{\textit{Test Set}}\\
            \hline
            $\text{BERT}_{base}$ &1.00$\times$ &52.1 &84.0 &93.5 &90.5 &88.9 &71.2 &66.4 &85.8 &79.1 \\
            \hline
		DPBERT(ours) &1.34$\times$ &53.5 &82.8 &91.9 &88.4 &86.5 &69.3 &65.9 &83.5 &77.7 \\
            \bottomrule
	\end{tabular}
	}
        \caption{Overall results on the GLUE benchmark. Results of development sets are averaged over 3 runs and we submit the model with the highest score to the leaderboard to obtain the results of test sets. We average the speed-up ratio across 9 tasks and obtain the overall speed-up ratio. The numbers under each dataset represent the number of training samples. We mark "-" on STS-B for DEEBERT since it do not support regression.}	
        
	\label{tab:overall}
\end{table*}

\subsubsection{Reinforcement Learning for the Entire Model} \label{train entire model}
Since {\tt DPBERT} make a series of discrete decisions at planning module, we use reinforcement learning to solve this process. In the context of policy estimation, we frame the task of estimating the planning function as follows. The planning policy is defined as a function taking input as the hidden state $\boldsymbol{h^i}$ and output the probability distribution over the planning action $a^i$, representing executing($a^i$=1) or bypassing($a^i$=0) the $i^{th}$ transformer layer:
\begin{equation}
    \pi(\boldsymbol{h^{i-1}})=\mathbb{P}(\text{Plan}^i(\boldsymbol{h^i}) = a^i)    
\end{equation}
where $\text{Plan}^i(i=1,2,...,L)$ is the set of policy networks that need to be updated in our proposal framework. We define the sequence of planning decision samples starting from input $x$ as:
\begin{equation}
    \boldsymbol{a}=[a^1, ..., a^L] \sim \pi_{\text{Trans}_\theta}(x)
\end{equation}
where $\text{Trans}_\theta=[\text{Trans}_\theta^1, ..., \text{Trans}_\theta^L]$ is the sequence of network layers parameterized by $\theta$ and ${a^i} \in \{0,1\}$. The planning reward $R^i$ corresponding to the planning network $\text{Plan}^i$ is defined as the future rewards:
\begin{equation}
    R^i = \frac{1}{L}\sum^{L-i}_{i=0}{(1-a^i)C^i} - \beta\mathcal{L}(\hat{y}(x,F_\theta, \boldsymbol{a},y))
\end{equation}
where the constant $C^i$ is the cost of executing $\text{Trans}^i$ and we set it to 1 in our experiment. $\mathcal{L}$ is the loss function corresponding to the downstream task. The adjustable hyperparameter $\beta$ can balance accuracy and model acceleration. 

In general, the greater the acceleration of the model, the worse the performance of the model, and vice versa. Thus, we need to constrain the degree of acceleration of the model. To control exactly the extent, We define a metric to estimate the number of layers the model skipped during inference:
\begin{equation}
    \mu = \frac{1}{L}\sum^L_{i=1}s_i
\end{equation}
 where $s_i$ is the softmax output of Equation \eqref{fc}. Given a target rate $t$, we impose an equality constraint $\mu=t$ by introducing a violation penalty:
 \begin{equation}
    \xi=(\mu - t)^2    
 \end{equation}
Introducing the penalty term into the gradient update process, we define the overall objective function as:
\begin{equation}
    \mathcal{J}(\theta,\pi)=\mathcal{L}(\hat{y}(x,F_\theta,\boldsymbol{a}\thicksim\pi(x),y) - \lambda_1 E_a\bigg[\sum^{L}_{i=1}{R^i}\bigg] + \lambda_2\xi    
    \label{rl_loss}
\end{equation}
The first component in Equation \eqref{rl_loss} is the task-specific function for which the gradient can be calculated directly, while the second component is the expected rewards for planning decisions whose gradients can be obtained by reinforcement learning. $\lambda_1,\lambda_2\in\mathbb{R}$ are two hyperparameters that balance the model acceleration and the skip rate.  

\section{Experiments}
In this section, we evaluate the effectiveness of {\tt DPBERT} on the GLUE benchmark with comparison to state-of-the-art baselines. 

\subsection{Datasets}
We evaluate our proposed approach on six classification datasets and one regression dataset of the GLUE benchmark  \cite{DBLP:conf/iclr/WangSMHLB19}. Specifically, we test on  Recognizing Textual Entailment (RTE)  \cite{DBLP:conf/iclr/WangSMHLB19}, Microsoft Research Paraphrase Matching(MRPC)  \cite{dolan2005automatically}, Multi-Genre Natural Language Inference Matched(MNLI-m), Multi-Genre Natural Language Inference Mismatched(MNLI-mm)  \cite{williams2017broad}, Question Natural Language Inference (QNLI)  \cite{rajpurkar2016squad}, Quora Question Pairs(QQP)\footnote{https://www.quora.com/q/quoradata/First-Quora-Dataset-Release-Question-Pairs} and Stanford Sentiment Treebank (SST-2)  \cite{socher2013recursive} for the classification task;  Semantic Textual Similarity Benchmark(STS-B)  \cite{conneau2018senteval} for the regression task. Note that we exclude WNLI \cite{levesque2012winograd} following the original BERT paper \cite{DBLP:conf/naacl/DevlinCLT19}.

For evaluating, we use accuracy  as the metric for RTE, SST-2, QNLI, MNLI-m and MNLI-mm. The average of accuracy and F1 are applied for QQP and MRPC. The results on STS-B are reported with the average of the Spearman and Pearson correlation. Mattew's correlation is used for CoLA.

\subsection{Baselines}
For the tasks mentioned above, we compare our method with three types of baselines and methods: 

\begin{itemize}
\item \textbf{Backbone models}: we choose 12-layer BERT-base model released by Google \cite{DBLP:conf/naacl/DevlinCLT19}. \item \textbf{Static approaches}: We report the performance of LayerDrop \cite{fan2019reducing}, an effective weight tuning method. For knowledge distillation, we include the results of DistilBERT \cite{sanh2019distilbert}. 
\item \textbf{Input-adaptive inference}: The early exiting methods including DEEBERT \cite{DBLP:conf/acl/XinTLYL20},  PABEE \cite{zhou2020bert} and BERxiT  \cite{DBLP:conf/acl/XinTLYL20} constraint the degree of speed-up through a threshold. 
\end{itemize}

Following the settings in the previous study \cite{zhou2020bert}, we search over a set of thresholds to find the one producing the best performance for the baselines. In this process, we constrain the speed-up ratio between 1.30$\times$ to 1.96$\times$ (the speed-up ratios of BERT-9L and -6L, respectively) to make a fair comparison. The results provided in some baselines are based on different datasets or models. For example,  DEEBERT only provides results for the test set; BERxiT’s comparison metric is the number of layers skipped instead of inference time; PABEE’s research are based on ALBERT. Therefore, we reproduce the results of these baselines without changing the hyperparameters in the original settings. 

\subsection{Experiment Setup}

\paragraph{Training} We added a sequence of fully-connected networks corresponding to the transformer layers as planning module. For fine-tuning and planning module initialization, we train separately 5 epochs with a learning rate of 2e-5 using AdamW \cite{kingma2014adam}. For reinforcement learning, we perform grid search over batch sizes of \{8, 16, 32\}, learning rates of \{2e-5, 3e-5, 4e-5, 5e-5\}, $\beta$ of \{5, 20, 35, 50\} and $\lambda_1$ of\{0.1, 0.5, 1, 1.5\} with an AdamW optimizer. We exploited a dynamic planning mechanism and selected the model with the best performance on the development set. We implemented {\tt DPBERT} on the base of Hugging Face's Transformers. We conducted our experiments on a single Nvidia 1080Ti 12GB GPU.

\paragraph{Inference} Following prior work \cite{DBLP:conf/acl/XinTLYL20,xin2021berxit}, we evaluate latency by performing inference on a per-instance basis. To simulate a common latency-sensitive production scenario when handling individual requests from various users, we set the batch size for inference as 1. The result we report is median performance over 3 runs with different random seeds. For {\tt DPBERT}, we set the target rate $t=0.4$ in the overall comparison to keep the speed-up ratio between 1.30$\times$ and 1.96$\times$ while obtaining good performance following table 2. We further analyse the behaviour of {\tt DPBERT} with different target rate settings in Section \ref{Accuracy-Latency Curve}.

\subsection{Comparison with Baselines}
The experimental results of our {\tt DPBERT} approach and baselines are shown in Table~\ref{tab:overall}. We first report the result on the development set. With comparison to baselines, we can see that our {\tt DPBERT} outperforms all baselines on the track of enhancing the inference efficiency of BERT, which verifies the effectiveness of our proposed approach. Due to the use of additional corpus (a concatenation of English Wikipedia and Toronto Book Corpus \cite{zhu2015aligning}) for pre-training as origin BERT, DistilBERT performs better on some data-sensitive tasks such as SST-2 and QNLI, while {\tt DPBERT} maintains an advantage   in the average score. For the Input-adaptive methods track, our approach outperforms DEEBERT on all tasks. BERxit and PABEE employs more sophisticated early-exit mechanisms to accelerate the inference, and further enhances the performance. However, {\tt DPBERT} still achieves better performance on the majority of tasks except QQP. Overall, our approach speeds up the inference of BERT by 1.34$\times$ while retaining 98\% accuracy.

To further demonstrate the effectiveness and the robustness of our approach, we submitted our model predictions to the official GLUE evaluation server\footnote{https://gluebenchmark.com/} to obtain results on the test set, as summarized in the second part of Table~\ref{tab:overall}. We can see that the performance of our approach is reduced by only 2\% compared to the backbone model. 

\begin{table}[t]
    \centering
    \begin{tabular}{lrrrr}
        \toprule
        MODEL  & RTE & MRPC & SST-2 & STS-B \\
        \midrule
        DPBERT &\textbf{66.8}  &\textbf{87.8} &\textbf{90.8} &\textbf{88.7} \\
        DPBERT-soft &51.6 &82.8 &85.7 &83.8 \\
        \bottomrule
    \end{tabular}
    \caption{Influence of reinforcement learning on the model accuracy. We denote by DPBERT-soft the scheme with soft output of planning network.}
    \label{tab:ablation}
\end{table}
\subsection{Perfomance-Latency Curve} \label{Accuracy-Latency Curve}

\begin{figure*}[!ht]
    \centering
    \includegraphics[scale=0.5]{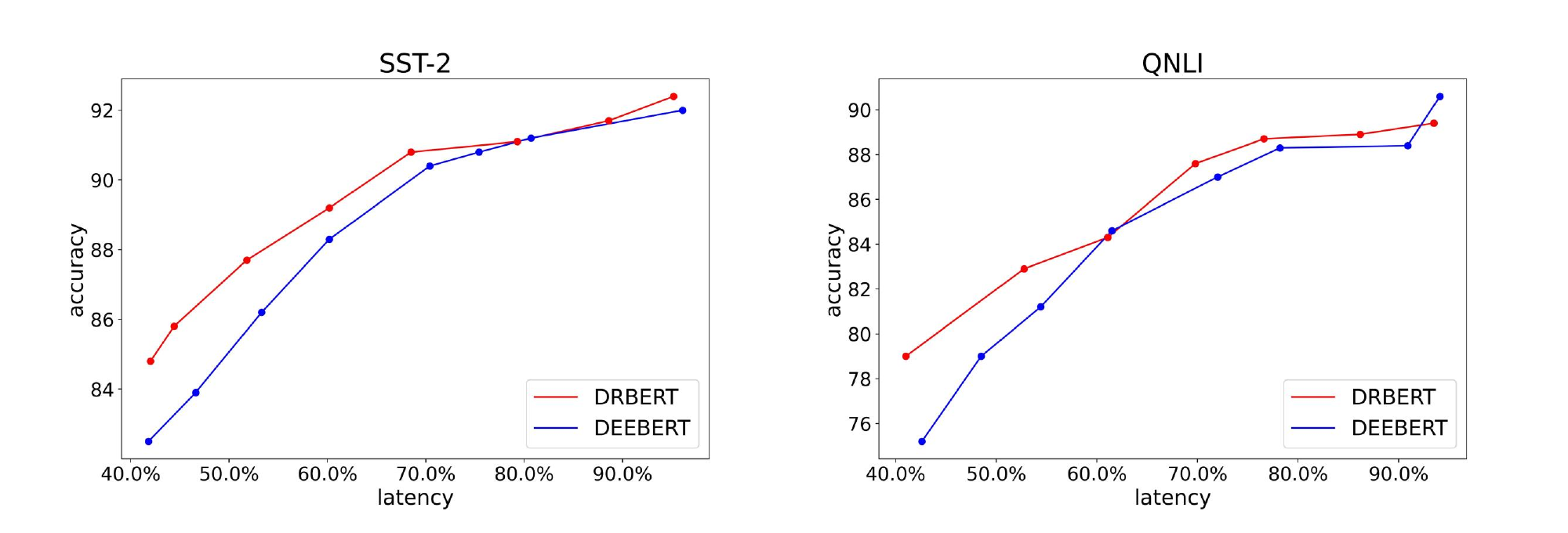}
    \caption{Performance-latency curve on the development set of two datasets in GLUE. The horizontal coordinate is the percentage of inference latency relative to the BERT-base.}\label{acc-latency curve}
\end{figure*}

In order to investigate the relationship between speed-up ratio and performance on {\tt DPBERT}, we explore the performance-latency curves by adjusting the target rate and early exiting threshold. The results generated on the development set of SST-2 and QNLI are shown in Figure \ref{acc-latency curve}. We can see that the model performance is inversely proportional to speed-up ratio. Comparing the results between DEEBERT and {\tt DPBERT} with respect to the trade-off between accuracy and inference time, we can see that {\tt DPBERT} outperforms DEEBERT on SST-2 consistently. On QNLI, DEEBERT yields slightly higher accuracy at low speed-up ratio. However, our {\tt DPBERT} outperforms the former in most cases. In addition, we can also see that {\tt DPBERT} produces a flatter curve on both tasks compared to DEEBERT, indicating that {\tt DPBERT} maintains better performance with high speed-up ratio compared to traditional early exiting.

\begin{figure*}[!ht]
    \centering
    \includegraphics[scale=0.65]{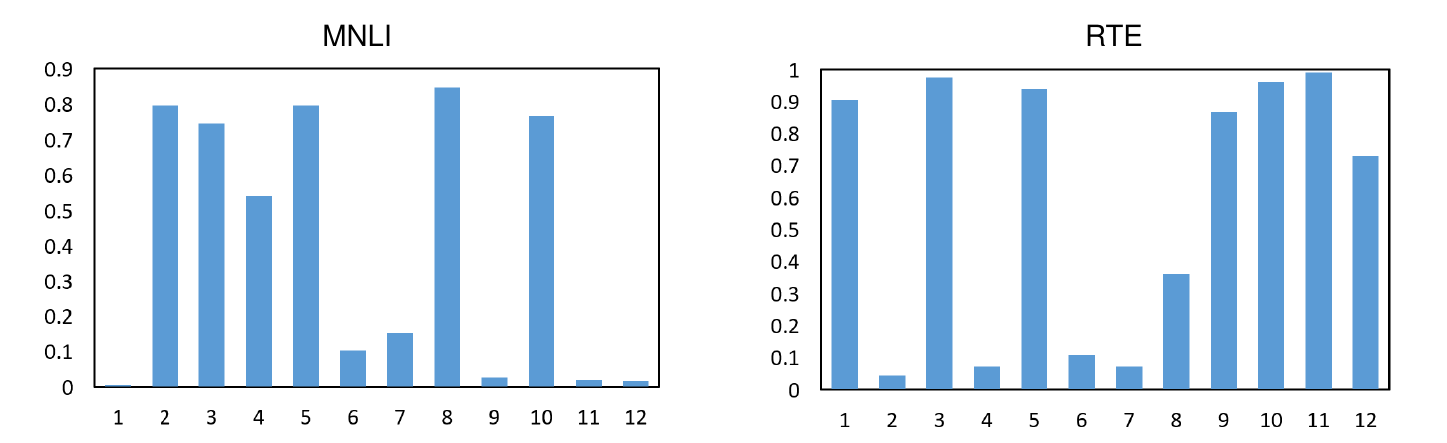}
    \caption{The distributions of layers executed during inference.}\label{layer selections}
\end{figure*}

\subsection{Selection of Computational Path}
In this section, we explored the specific implementation of the dynamic planning mechanism. In the previous section, we mentioned that dynamic planning provides more computational path choices relative to early exiting, which facilitates finding the optimal trade-off between model speed-up and performance.

To further demonstrate the versatility brought by dynamic planning, we calculated the frequency of different layers being chosen across datasets and result is shown in Figure \ref{layer selections}. From the distribution we can observe that our proposal focuses more on lexical and syntax information in bottom and middle layers on MNLI, while it focuses on using more top layers to capture semantic knowledge in RTE.It illustrates that that our approach could choose computational paths more freely according to different tasks as well as samples (Section \ref{Case study}) as opposed to the early exiting model that can only skip the top layer.
\begin{table*}[!ht]
  \centering
    \begin{tabular}{l|c|l} 
      \toprule
      Dataset & Example & layers\\
      \midrule
      \multirow{3}{*}{MNLI} & \makecell[l]{Premise: You and your friends are not welcome here, said Severn. \\ 
      Hypothesis: Severn said the people were not welcome there. \\  
      Label: entailment} & 1 2 3 4 5 6\\
      \cline{2-3}
      & \makecell[l]{Premise: Other villages are much less developed, and therein lies the essence of \\ many delights.\\	
      Hypothesis: If more people lived in the villages the development would skyrocket.	\\
      Label: neutral} & 1 2 3 4 5 6 8 10 11 12\\
      \hline
      \multirow{2}{*}{QNLI} & \makecell[l]{Question: What is the hottest temperature record for Fresno? \\
      Context: Measurable precipitation falls on an average of 48 days annually. \\
      Label: not entialment} & 1 3 4 5\\
      \cline{2-3}
      & \makecell[l]{Question: Agassiz's approach to science combined observation and what? \\
      Context: When it came to explaining life-forms, Agassiz resorted to matters \\ of shape based on a presumed archetype for his evidence. \\
      Label: not entialment} & 1 3 4 5 9 10 12\\
      \bottomrule
    \end{tabular}
  \caption{Case study of the dynamic planning. }
  \label{tab:case}
\end{table*}

\subsection{Ablation Study} \label{Ablation Study}
To investigate the effectiveness of reinforcement learning, We conducted ablation study through changing the training process and objective function. Before the joint training, We kept fine-tuning the backbone and initialization for the planning module. In the third stage (Section \ref{train entire model}), we relaxed the output of the planning network to a continuous range of [0,1], similar to the second stage (Section \ref{intialization}). Additionally, the second component about reinforcement learning was removed in Equation \eqref{rl_loss} while training the whole model. We denote this method by DPBERT-soft. We report the results on the development set of four tasks in GLUE benchmark. 

The results of two methods are shown in Table \ref{tab:ablation}. We adjusted the target rate such that the speed-up ratio is the same for both methods. From the table, we can see that the performance decreases when we remove the module corresponding to reinforcement learning, which indicates that our reinforcement learning framework is effective for the model to learn the policy for planning. In order to find the subsequence of transformer layers that yields the best accuracy-speed trade-off, the planning module of the model needs to make a sequence of decision actions. Compared to supervised learning, our reinforcement learning framework that considers long-term reward is better suited to finding the optimal solution to sequential decision problems, indicating that our approach with the planning module based on reinforcement learning performs more effectively.

\subsection{Case Study} \label{Case study}

In this section, we investigate the characteristics of dynamic planning through conducting a particular case study on various datasets in GLUE benchmark. As show in Table \ref{tab:case}, we present a variety of examples and their corresponding layers included in computational graph during inference in the development set of MNLI and QNLI. As summarized in the previous work\cite{jawahar2019does}, BERT embeds a rich hierarchy of linguistic signals: surface information at the bottom, syntactic information in the middle, semantic information at the top, which means that the functions of computational graphs made up of different layer combinations are different. We explore whether {\tt DPBERT} matches the computational graph with the appropriate functionality to the characteristics of the example. 

In the first example about MNLI, a task about sentence pair relational reasoning, we can observe that the hypothesis simply changes the order of words compared to the premise. The relation of the sentence pair can be inferred depending on surface information and  simple syntactic knowledge. Therefore, our approach selects the first six layers to be the computational path. Compared with the former, the second example is obviously more complicated. The keywords in the premise and hypothesis are similar, such as "villages" and "develop". However, the apparent content of their expressions varies considerably. To understand their semantic information, {\tt DPBERT} increases the layers at the middle and top. The second part of Table \ref{tab:case} presents the examples of QNLI, a task aims to predict whether the context contains the answer to the question. As shown in the first example, there is rare overlap of lexical information of the question and answer, so it is easy to analyze that this is a negative example. Therefore, our method executes only four layers at the bottom for it while including more top layers in the computational graph for the second example, which presents question and context with similar keywords thus requiring further analysis.

Through the examples above, we can see that {\tt DPBERT} tends to assign the bottom layer to the superficial example while executing more high layers for uncertain samples. It demonstrated that our approach can construct the computational path based on the difficulty of sample during inference.

\section{Conclusion}
In this paper, we presented {\tt DPBERT}, a straightforward but
effective approach to select a subsequence of transformer layers list of backbone as the computational graph for each sample. Specifically, {\tt DPBERT} accomplishes this by adding a sequence of planning network corresponding to the transformer layer to decide whether a layer is included in the path during inference. To find the optimal trade-off of model speed-up and accuracy, we introduce reinforcement learning in the training framework and update the parameters of planning module based on policy optimization. Our experiments show significant results on nine datasets on GLUE benchmark. Empirical results have demonstrated that {\tt DPBERT} could achieve 98\% performance of BERT while significant decreasing inference time. Compared to prior input-adaptive techniques, it offers significantly better trade-off between accuracy and inference time. Furthermore, we demonstrated that dynamic planning brings more versatility compared to early exiting i.e. provides more computational path choices. For future work, we plan to extend {\tt DPBERT} to other variant model of BERT (e.g., ALBERT) and try to apply our approach to the models in natural language generation domain to explore whether it works on this task. in addition, integrating our input-adaptive technique with static methods (e.g., knowledge distillation) might potentially result in higher efficiency, which we intend to investigate in the future. It would also be interesting to investigate the integration of symbolic planning model learning \cite{DBLP:conf/aips/ZhuoYPL11,DBLP:journals/ai/Zhuo014,DBLP:journals/ai/ZhuoK17,DBLP:conf/aaai/JinMJZCY22} into {\tt DPBERT} to help improve the explainability of dynamic planning. 

\bibliography{ecai}

\end{document}